\documentclass[conference]{IEEEtran}
\IEEEoverridecommandlockouts
\usepackage{amsmath,amssymb,amsfonts}
\usepackage{algorithmic}
\usepackage{graphicx}
\usepackage{textcomp}
\usepackage{biblatex}
\usepackage{booktabs}
\usepackage{tabularx}
\usepackage{array}

\usepackage{xcolor}
\def\BibTeX{{\rm B\kern-.05em{\sc i\kern-.025em b}\kern-.08em
    T\kern-.1667em\lower.7ex\hbox{E}\kern-.125emX}}
\begin{document}

\title{Harnessing FPGA Technology for Enhanced Biomedical Computation\\
\thanks{*Corresponding author}
}

\makeatletter
\newcommand{\linebreakand}{%
  \end{@IEEEauthorhalign}
  \hfill\mbox{}\par
  \mbox{}\hfill\begin{@IEEEauthorhalign}
}
\makeatother

\author{

\IEEEauthorblockN{Nisanur Alici*}
  \IEEEauthorblockA{\textit{Biomedical Eng.} \\
    \textit{Erciyes University}\\
    Kayseri, Turkey \\
    alicinisanur@gmail.com}
  \and 

\IEEEauthorblockN{Kayode Inadagbo}
  \IEEEauthorblockA{\textit{Electrical and Computer Eng.} \\
    \textit{Prairie View A\&M University}\\
    Prairie View, TX \\
    kinadagbo@pvamu.edu}
  \and
  
  \IEEEauthorblockN{Murat Isik}
  \IEEEauthorblockA{\textit{Electrical and Computer Eng.} \\
    \textit{Drexel University}\\
    Philadelphia, PA \\
    mci38@drexel.edu}
}

\IEEEoverridecommandlockouts
\maketitle
\IEEEpubidadjcol
\begin{abstract}
This research delves into sophisticated neural network frameworks like Convolutional Neural Networks (CNN), Recurrent Neural Networks (RNN), Long Short-Term Memory Networks (LSTMs), and Deep Belief Networks (DBNs) for improved analysis of ECG signals via Field Programmable Gate Arrays (FPGAs). The MIT-BIH Arrhythmia Database serves as the foundation for training and evaluating our models, with added Gaussian noise to heighten the algorithms' resilience. The developed architectures incorporate various layers for specific processing and categorization functions, employing strategies such as the EarlyStopping callback and Dropout layer to prevent overfitting. Additionally, this paper details the creation of a tailored Tensor Compute Unit (TCU) accelerator for the PYNQ Z1 platform. It provides a thorough methodology for implementing FPGA-based machine learning, encompassing the configuration of the Tensil toolchain in Docker, selection of architectures, PS-PL configuration, and the compilation and deployment of models. By evaluating performance indicators like latency and throughput, we showcase the efficacy of FPGAs in advanced biomedical computing. This study ultimately serves as a comprehensive guide to optimizing neural network operations on FPGAs across various fields.

\end{abstract}

\section{Introduction}
High-performance computing systems are essential for real-time and accurate biomedical applications, such as ECG signal analysis. Field-Programmable Gate Arrays (FPGAs), known for their unique combination of flexibility, performance, and energy efficiency, are increasingly favored for accelerating computations across various domains, including biomedical engineering. This paper focuses on efficiently implementing Convolutional Neural Networks (CNNs), Recurrent Neural Networks (RNNs), Long Short-Term Memory Networks (LSTMs), and Deep Belief Networks (DBNs) on FPGAs for tasks like arrhythmia detection, heartbeat classification, and risk stratification in electrocardiogram (ECG) signal analysis.

FPGA-based accelerators offer several advantages in biomedical applications. They provide high parallelism and are crucial for large-scale data processing due to their ability to handle multiple tasks simultaneously. Their customizable architecture enables performance optimization for specific applications, a feat unachievable with general-purpose processors. Another significant advantage of FPGAs is their low latency and high bandwidth, making them ideal for real-time processing and large data transfers. Additionally, their energy efficiency renders them ideal for power-intensive applications like image processing, machine learning, and real-time processing \cite{isik2023energy, 10274450}.

As the demand for biomedical applications grows, the need for high-performance computing systems capable of processing large data volumes quickly and accurately becomes increasingly crucial. FPGA-based accelerators, with their capacity to meet these demands effortlessly, pave the way for significant advancements in biomedical engineering, especially with the efficient implementation of CNNs, RNNs, LSTMs, and DBNs for ECG signal analysis.

The rest of the paper is organized as follows: \textbf{Section II} presents the motivation behind our work and a review of related studies in the field. \textbf{Section III} introduces open-source machine learning (ML) inference accelerators, with a focus on the Tensil AI accelerator. In \textbf{Section IV}, we describe our proposed method for efficient FPGA implementation of CNNs, RNNs, LSTMs, and DBNs for ECG signal analysis. \textbf{Section V} presents the experimental results and provides an analysis of the performance and efficiency of our proposed method. Finally, \textbf{Section VI} concludes the paper, summarizing our contributions and discussing future directions for research in this area.

\section{Motivation}
The relentless pursuit of computational efficiency in high-performance computing (HPC) has propelled industries toward rapid innovation and enhanced problem-solving capabilities. Among the myriad technologies that have emerged, FPGAs stand out as a transformative force in this evolution, especially in applications that demand real-time processing and analysis. The motivation behind integrating FPGAs into HPC systems stems from their unique ability to be reconfigured for specific tasks, offering unparalleled flexibility and performance benefits compared to traditional CPUs and GPUs \cite{EGILA2016513, isik2023astrocyte}.

In the biomedical field, the impact of this technology is particularly palpable. FPGAs can drastically improve the processing of ECG signals, which are crucial for monitoring and diagnosing cardiac conditions. The parallel processing capabilities of FPGAs, coupled with their reconfigurable nature, allow for the design of optimized systems that process ECG signals much more rapidly than conventional computing systems. This rapid analysis is critical in medical emergencies, such as in the detection of life-threatening conditions like arrhythmia or cardiac ischemia \cite{electronics10192324}.

Combining FPGAs with deep learning techniques has advanced ECG signal analysis substantially. Deep learning algorithms are particularly effective due to their inherent ability to learn and adapt from data. For example, Convolutional Neural Networks (CNNs) can automatically learn spatial hierarchies of features, while Recurrent Neural Networks (RNNs) and Long Short-Term Memory networks (LSTMs) excel in processing sequential data. FPGAs facilitate the implementation of these advanced algorithms in a power and cost-efficient manner \cite{Kumar2023}. CNNs have proven highly effective in arrhythmia classification and detection, exploiting their ability to capture spatial dependencies in ECG data. Studies have demonstrated the successful implementation of CNNs on FPGAs, achieving high accuracy in real-time patient monitoring systems. RNNs and LSTMs extend this capability by processing the temporal characteristics of ECG signals, where the sequence and timing of heartbeats provide insights into cardiac rhythm and identify abnormalities \cite{Kumar2023} \cite{Liu2023-sx}. Furthermore, the integration of FPGAs in ECG signal analysis has facilitated the development of modern implantable cardiac devices such as pacemakers, where their high-performance and energy-efficient characteristics are crucial \cite{Liu2023-rz}.

Beyond ECG signal analysis, FPGAs find applications in various other biomedical domains. In medical imaging, for example, FPGAs are crucial for processing large datasets in real-time, as required in MRI, CT scans, and ultrasound devices \cite{jimaging8040114}. Minimally invasive surgery platforms benefit from FPGAs' computational power and real-time processing capabilities, critical for precision and control in these procedures \cite{Nagyne_Elek2022-vh}. Digital image processing techniques like inverting image operations, brightness control, segmentation, and contrast stretching in biomedical applications are enhanced by FPGAs for improved image quality and aid in diagnosis \cite{jimaging8040114}. FPGAs also accelerate real-time volume rendering for 3D medical images, using techniques such as block-based ray casting to efficiently process and visualize complex volumetric data \cite{5639475}.

Apart from the biomedical domain, FPGAs have applications in image and video processing, where they handle high-resolution data in real-time. FPGA-based image processing systems have achieved significant throughput, processing up to 52 frames per second in filtering applications and 20 frames per second in image segmentation tasks, showcasing FPGAs' potential in multimedia applications \cite{DBLP:journals/corr/abs-1710-05154}. In the financial industry, FPGAs are used for complex mathematical operations in high-frequency trading systems and financial instrument calculations, such as options and futures, achieving processing times of less than one microsecond \cite{kohda2021characteristics, kohda2022characteristics}. FPGAs have also emerged as a vital component in the Internet of Things (IoT) domain, particularly in edge computing for real-time data analysis and decision-making processes. Their reconfigurability enables adaptation to the dynamic IoT landscape, facilitating efficient data traffic management, robust security protocols, and deployment of machine learning algorithms in IoT gateways \cite{216801}.

Despite their challenges and limitations, such as limited on-chip memory, floating-point support, and limited availability of prebuilt IP blocks, the advantages of FPGAs have made them increasingly popular in high-performance computing applications. Recent developments in high-level synthesis tools and the availability of pre-built IP blocks have made FPGAs more accessible for high-performance computing. As development tools, IP blocks, and FPGA-based solutions evolve, the adoption of FPGAs in high-performance computing, particularly in biomedical applications like ECG signal analysis, will increase \cite{huang2019accelerating}.

\section{Open-source ML inference accelerators}
Machine learning (ML) inference plays a crucial role in numerous high-performance computing applications. This process involves using models to analyze input data and generate output results. High-performance computing systems are instrumental in accelerating ML inference, a task that is often computationally intensive. ML inference accelerators that are open-source may be beneficial to high-performance computing applications. A machine learning inference accelerator is a specialized hardware device designed to efficiently execute ML inference tasks. Traditionally, ML inference tasks are performed on general-purpose processors or graphics processing units (GPUs), which are not specifically optimized for machine learning. A machine learning inference accelerator streamlines and optimizes the execution of machine learning inference tasks. A major advantage of open-source ML inference accelerators is that they are free and can be customized for specific use cases. They offer a transparent development process that encourages community participation. Furthermore, open-source accelerators can reduce the cost and time associated with the development of ML inference accelerators. The Versatile Tensor Accelerator (VTA), an open-source ML inference accelerator, has recently garnered significant attention. An optimized hardware accelerator is used for performing inference tasks using VTA. VTA supports TensorFlow, PyTorch, and ONNX among other ML frameworks. VTA is compatible with various hardware platforms, including FPGAs and ASICs \cite{moreau2019hardware}.

\begin{figure}[h!]
    \centering
    \includegraphics[width=0.40\textwidth]{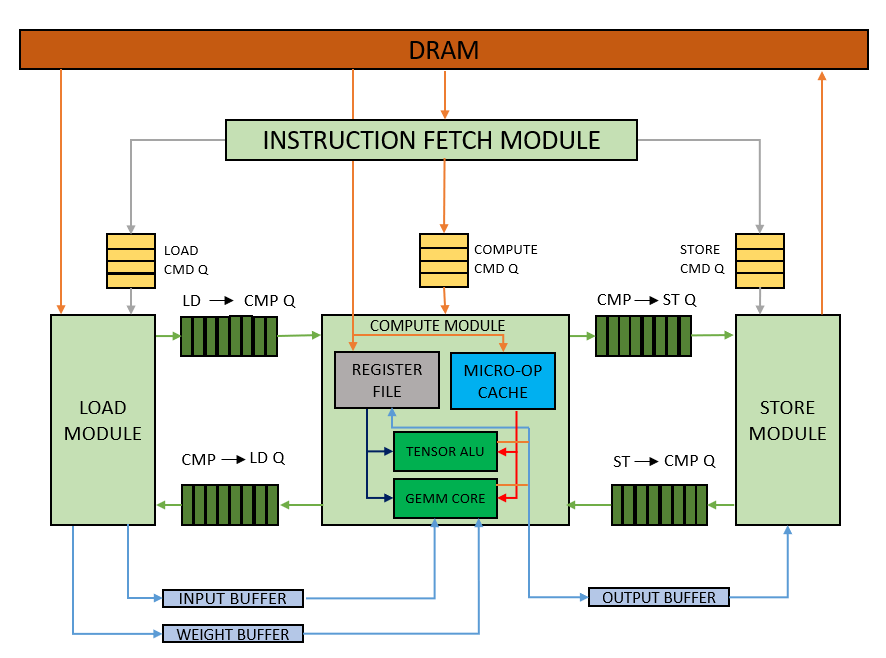}
    \caption{VTA Framework}\cite{moreau2019hardware}
    \label{fig_7}
\end{figure}

The availability of open-source tools and platforms allows developers to collaborate to develop better ML inference solutions. In addition to VTA \cite{zunin2021intel} \cite{WinNT2}, other open-source ML inference accelerators are available, such as Intel's OpenVINO and Xilinx's Deep Learning Processor. These accelerators provide a variety of options for the development and optimization of machine learning systems. Open-source ML inference accelerators are flexible and powerful tools that can be used in high-performance computing applications. They enable the customized and optimized development of ML inference systems, reducing development costs and time, fostering collaboration and innovation, and leading to the wide adoption of ML techniques. Open-source ML inference accelerators will continue to become more powerful and efficient as the field of ML grows and evolves. Frameworks such as Nengo and Tensil AI enable the development of high-performance computing applications utilizing FPGAs. Nengo software can be used to build large-scale neural models on a variety of hardware platforms, including FPGAs. Its flexibility and extensibility allow users to create customized algorithms and models. Nengo is particularly suited for applications in robotics and cognitive modeling due to its capability to construct complex models with numerous neurons and synapses. The Tensil AI hardware accelerator performs machine learning inference tasks. Tensil AI offers high performance at low power consumption, making it ideal for applications such as image recognition and natural language processing. Tensil AI supports a variety of machine learning frameworks, including TensorFlow and PyTorch, and can easily be integrated with existing hardware architectures. One major difference between Tensil AI and Nengo is their focus. Tensil AI aims to accelerate machine learning inference, while Nengo is geared towards building large-scale neural networks \cite{morcos2019nengofpga} \cite{dewolf2020nengo} \cite{gosmann2017automatic} \cite{isik2023design}.  

\vspace{3pt}Nengo can be used for a wider range of tasks compared to the more specific Tensil AI. The development processes of Nengo and Tensil AI also differ fundamentally. Nengo is an open-source project actively developed and maintained by a large developer community, providing a variety of resources such as documentation, tutorials, and support forums. Tensil AI, on the other hand, is a commercial product developed and supported by its company, offering dedicated support and resources, but lacking the extensive community support available for open-source software. Tensil AI efficiently performs inference tasks in machine learning. For instance, it enables rapid inferences in self-driving cars and industrial automation. Nengo, on the other hand, simulates complex behaviors over long periods using large-scale neural models. Tensil AI's potential drawback is its limited flexibility due to the specialized nature of its hardware accelerator. Users may not be able to customize models and algorithms as extensively and may have to rely on prebuilt models and architectures. Both Nengo and Tensil AI are powerful frameworks for developing high-performance computing applications, with Tensil AI being more suited for specific tasks like machine learning inference, while Nengo caters to a wider range of applications. Developers should carefully evaluate the strengths and weaknesses of each framework before deciding, as the ultimate choice depends on their specific application needs \cite{WinNT} \cite{WinNT1}.

\section{Method}
Our framework assesses the performance and power profiles of ML algorithms across three dimensions: i) applications, ii) algorithms, and iii) hardware architectures, as depicted in \ref{Figure 3}. Within the application dimension, we categorize tasks into computer vision, natural language processing, and time-series data analysis. These applications are implemented using four distinct ML algorithms: Long Short-Term Memory (LSTM), Convolutional Neural Networks (CNN), Recurrent Neural Network (RNN), and Deep Belief Network (DBN), each tested on different hardware platforms including CPU, GPU, and FPGA. We classify energy estimation approaches into two types: top-down and bottom-up techniques. The bottom-up method utilizes a detailed statistical model that encompasses both workload analysis and measured power consumption, while the top-down approach estimates power usage based on the observed performance during algorithm simulations.

\begin{figure*}
    \graphicspath{ {D:\Stack} }
    \center \includegraphics[width=0.6\textwidth]{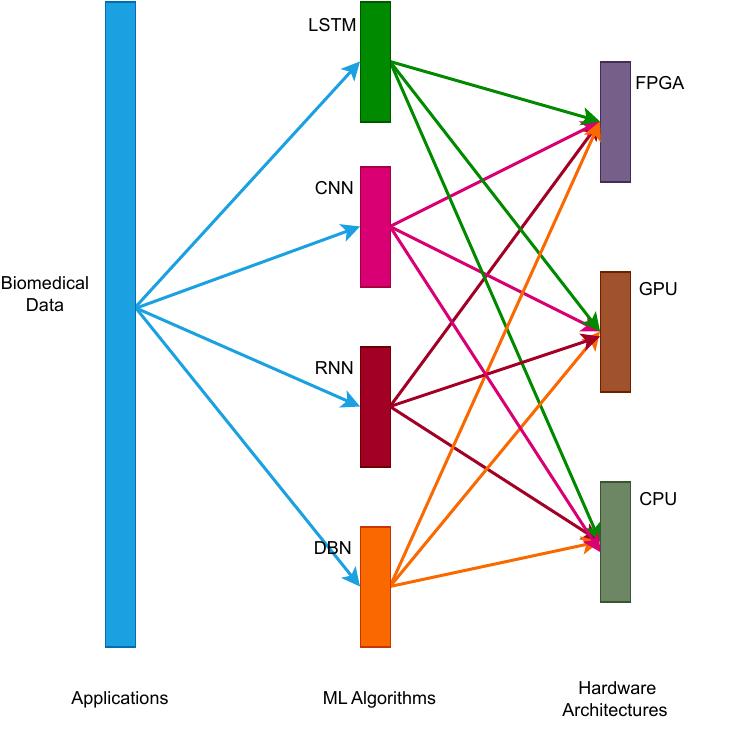}
    \caption{Biomedical data simulated with four machine learning algorithms realized on three types of hardware architectures. Note: Long Short-Term Memory (LSTM), Convolutional Neural Networks (CNN), Recurrent Neural Network (RNN), and Deep Belief Network (DBN).}
    \label{Figure 3}
\end{figure*}

\subsection{Dataset}

The MIT-BIH Arrhythmia Database offers valuable cardiac arrhythmia data through ECG recordings. Between 1975 and 1979, 48 half-hour excerpts of two-channel ambulatory ECG recordings were obtained from 47 subjects at Boston's Beth Israel Hospital. These recordings were digitized at a rate of 360 samples per second per channel, with 11-bit resolution over a 10 mV range. A total of 80\% of the dataset is devoted to training, while 20\% of the dataset is for validation. The original training data includes approximately 110,000 beats, each annotated by at least two cardiologists.

In contrast to the MNIST digit dataset, the MIT-BIH Arrhythmia Database does not inherently contain noisy variations. However, for specific applications, one can add noise to the ECG recordings. For example, to simulate real-world noisy conditions in ECG data, random Gaussian noise can be applied to the dataset elements. Similar to white noise, Gaussian noise follows a Gaussian distribution. This addition results in impulses with random values.

Using the modified dataset, now containing Gaussian noise, deep-learning models can be trained for ECG signal analysis, arrhythmia detection, and algorithm robustness improvement in noisy environments. Our dataset is shown in Fig. \ref{Figure 2}.
\begin{figure}[h]
    \graphicspath{ {D:\Stack} }
    \center \includegraphics[width=8cm, height=4cm]{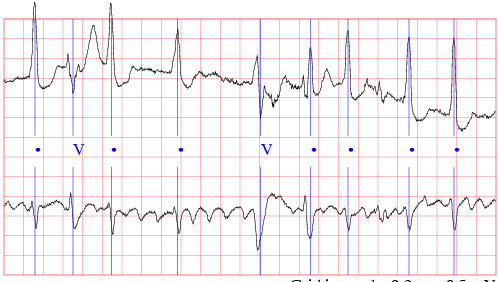}
    \caption{Project dataset}
    \label{Figure 2}
\end{figure}

\subsection{Training Details}

Our approach utilizes a four-layer neural network incorporating LSTM for ECG signal classification. The LSTM layers understand long-term data dependencies, while dense layers assist in classification. To prevent overfitting, a dropout layer randomly disables certain weights, and an EarlyStopping callback halts training once the validation loss stabilizes. There are 64 and 32 LSTM cells in the first and second LSTM layers, respectively. The dense layer carries out classification, producing five outputs corresponding to five classes. During compilation, a loss function and the 'man' optimizer are utilized for training. 'Accuracy' measures performance, and an EarlyStopping callback prevents overfitting. Training results are visualized with validation data, and the model's efficiency is evaluated using latency and throughput metrics.

The code comprises a neural network featuring several layers, including a BernoulliRBM for high-level abstractions, LogisticRegression for classification, and Dropout to prevent overfitting. The Flatten layer transforms 3D input into 1D, the Reshape layer adjusts dimensions, and the StandardScaler ensures data consistency. After data checks, the class imbalance is tackled with SMOTE. Deep Belief Networks (DBNs) are used on training data. The DBN model is evaluated using different sets. A CNN model classifies heartbeats in this code. Following data preprocessing and renaming of columns, the target variable distribution is plotted. Class imbalances are identified and balanced via SMOTE. The CNN model consists of Conv1D, MaxPool1D, and Dropout layers. Output is passed to a dense layer after flattening. This code provides a solid foundation for heartbeat classification using CNNs, but it can be further enhanced with detailed comments, dataset information, preprocessing steps, and hyperparameter tuning. The final part employs an RNN with a Bidirectional LSTM for ECG signal classification. The MIT-BIH Arrhythmia dataset is processed and split into training and validation sets. Model performance is assessed using latency, throughput, and other metrics.

For ECG analysis, the ideal model is one that has low latency and high throughput. These models can be compared based on their characteristics as follows:

\begin{itemize}
    \item \textbf{LSTM (Long Short-Term Memory):} LSTM is a type of recurrent neural network (RNN) architecture. It has gained significant prominence in the field of deep learning and artificial intelligence due to its ability to effectively model and capture long-range dependencies in sequential data. LSTM is a powerful tool in applications such as natural language processing, speech recognition, and time series forecasting.

\begin{figure}[h]
    \graphicspath{ {D:\Stack} }
    \center \includegraphics[width=8cm, height=4cm]{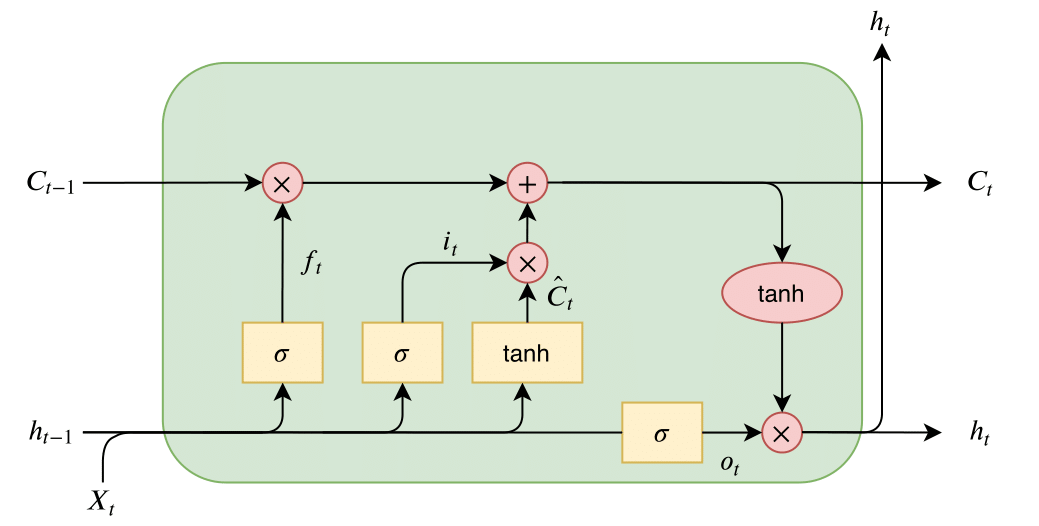}
    \caption{LSTM Architecture}
    \label{fig:lstm_architecture}
\end{figure}   
\cite{LSTMarchitecture}

\section*{LSTM Mathematical Model}

LSTMs have three gates:

1. \textbf{Forget Gate (\(f_t\))}: Determines what information from the previous cell state should be discarded or retained.

\begin{equation}
f_t = \sigma(W_f \cdot [h_{t-1}, x_t] + b_f)
\end{equation}

Where:
\begin{align*}
f_t & \text{ is the forget gate activation vector.} \\
\sigma & \text{ represents the sigmoid activation function.} \\
W_f & \text{ is the weight matrix for the forget gate.} \\
h_{t-1} & \text{ is the previous cell state.} \\
x_t & \text{ is the current input at time } t. \\
b_f & \text{ is the bias for the forget gate.}
\end{align*}

2. \textbf{Input Gate (\(i_t\))}: Determines which values from the input and the previous cell state should be updated.

 \begin{equation}
i_t = \sigma(W_i \cdot [h_{t-1}, x_t] + b_i)
\end{equation}

Where:
\begin{align*}
i_t & \text{ is the input gate activation vector.} \\
\sigma & \text{ represents the sigmoid activation function.} \\
W_i & \text{ is the weight matrix for the input gate.} \\
h_{t-1} & \text{ is the previous cell state.} \\
x_t & \text{ is the current input at time } t. \\
b_i & \text{ is the bias for the input gate.}
\end{align*}

3. \textbf{Cell State Update (\(\tilde{C}_t\))}: Combines new information from the current input and the previous cell state.

\begin{equation}
\tilde{C}_t = \tanh(W_C \cdot [h_{t-1}, x_t] + b_C)
\end{equation}

Where:
\begin{align*}
\tilde{C}_t & \text{ is the candidate cell state.} \\
\tanh & \text{ represents the hyperbolic tangent activation function.} \\
W_C & \text{ is the weight matrix for the cell state update.} \\
h_{t-1} & \text{ is the previous cell state.} \\
x_t & \text{ is the current input at time } t. \\
b_C & \text{ is the bias for the cell state update.}
\end{align*}

4. \textbf{Update and Output Gate (\(o_t\))}: Determines the new cell state and the output at the current time step.

 \begin{equation}
o_t = \sigma(W_o \cdot [h_{t-1}, x_t] + b_o)
\end{equation}

\begin{equation}
C_t = f_t \cdot C_{t-1} + i_t \cdot \tilde{C}_t
\end{equation}

\begin{equation}
h_t = o_t \cdot \tanh(C_t)
\end{equation}

Where:
\begin{align*}
o_t & \text{ is the output gate activation vector.} \\
C_t & \text{ is the new cell state.} \\
h_t & \text{ is the output vector for the current time step.} \\
C_{t-1} & \text{ is the previous cell state.} \\
\sigma & \text{ represents the sigmoid activation function.} \\
\tanh & \text{ represents the hyperbolic tangent activation function.} \\
W_o & \text{ is the weight matrix for the output gate.} \\
h_{t-1} & \text{ is the previous cell state.} \\
x_t & \text{ is the current input at time } t. \\
b_o & \text{ is the bias for the output gate.}
\end{align*}

LSTMs are effective for modeling sequences and time-series data, capturing long-range dependencies and storing information over extended sequences. They are designed to address the limitations of traditional RNNs, such as the vanishing and exploding gradient problems. ECG analysis can benefit from LSTMs when dealing with sequential data that exhibit long-term patterns. However, due to their complex structure, LSTMs may have higher latency and lower throughput compared to CNNs.

\item \textbf{CNN (Convolutional Neural Network):} CNN is a deep learning model revolutionizing the field of computer vision. It is specially designed for processing and analyzing visual data, making it ideal for tasks such as image recognition, object detection, and video analysis. CNNs consist of multiple layers, each designed for a specific purpose:

    1. Convolution Layers: These layers learn features automatically from the input data. They apply filters to detect various patterns, edges, and textures, enabling CNNs to learn hierarchical representations of visual features.
    
    2. Pooling Layers: These layers downsample feature maps from convolutional layers, reducing spatial dimensions and making the model robust to scale and position variations.
    
    3. Fully Connected Layers: These layers make predictions based on learned features. They classify or regress based on the high-level features extracted by previous layers.

\begin{figure}[h]
    \graphicspath{ {D:\Stack} }
    \center \includegraphics[width=8cm, height=4cm]{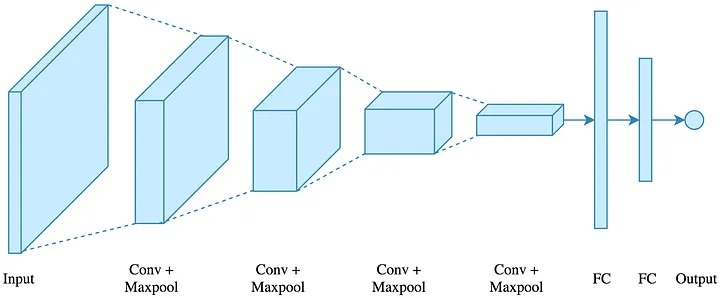}
    \caption{CNN Architecture}
    \label{fig:cnn_architecture}
\end{figure}   
\cite{CNNarchitecture}
\section*{CNN Mathematical Model}

1. \textbf{Convolution Operation (Convolutional Layer)}:

CNNs apply a set of learnable filters (kernels) to the input data through convolution operations. These filters convolve across the input to produce feature maps. The mathematical operation is represented as:

\begin{equation}
S(i, j) = (I * K)(i, j) = \sum_m \sum_n I(i - m, j - n) \cdot K(m, n)
\end{equation}

Where:
\begin{align*}
S(i, j) & \text{ represents the value at position } (i, j) \text{ in the feature map.} \\
I & \text{ denotes the input data.} \\
K & \text{ is the convolutional kernel (filter).} \\
(i, j) & \text{ are the spatial coordinates of the output feature map.} \\
(m, n) & \text{ are the coordinates within the kernel.}
\end{align*}

2. \textbf{Pooling Layer}:

Pooling layers, typically used after convolution, reduce the spatial dimensions of feature maps. Max pooling, which selects the maximum value from a local region of the input, is a common operation:

\begin{equation}
O(i, j) = \max_{m, n} I(i \cdot s + m, j \cdot s + n)
\end{equation}

Where:
\begin{align*}
O(i, j) & \text{ is the value at position } (i, j) \text{ in the pooled feature map.} \\
I & \text{ refers to the input feature map.} \\
s & \text{ is the stride, determining the step size for pooling.}
\end{align*}

3. \textbf{Fully Connected Layer}:

CNNs typically include one or more fully connected layers after convolutional and pooling layers to perform classification or regression. These layers function like traditional feedforward neural network layers:

\begin{equation}
y = \sigma(Wx + b)
\end{equation}

Where:
\begin{align*}
y & \text{ is the layer's output.} \\
\sigma & \text{ denotes the activation function (usually ReLU or sigmoid).} \\
W & \text{ is the weight matrix.} \\
x & \text{ is the input vector.} \\
b & \text{ is the bias vector.}
\end{align*}

Due to their ability to capture local patterns in data, CNNs excel in image and signal processing tasks, offering lower latency and higher throughput than LSTMs and RNNs for real-time applications.

\item \textbf{RNN (Recurrent Neural Network):}  RNN, which stands for Recurrent Neural Network, is a class of neural networks particularly well-suited for tasks that involve sequential data. Unlike traditional feedforward neural networks, RNNs have a unique ability to capture patterns and dependencies in data sequences, making them indispensable in a wide range of applications, including natural language processing, speech recognition, and time series analysis.

\begin{figure}[h]
    \graphicspath{ {D:\Stack} }
    \center \includegraphics[width=8cm, height=4cm]{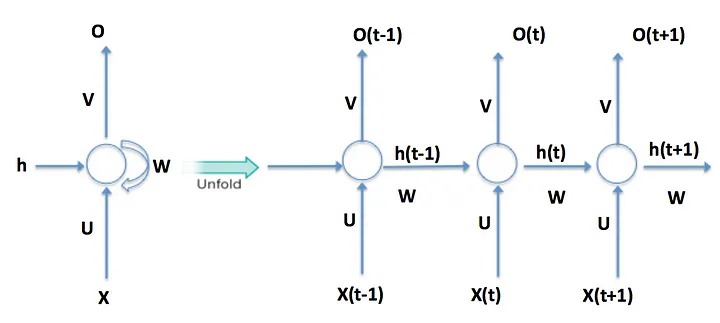}
    \caption{RNN Architecture}
    \label{fig:enter-label}
\end{figure}   
\cite{RNNarchitecture}

\section*{RNN Mathematical Model}

1. \textbf{Hidden State Update}:

Each time step \(t\) in an RNN updates a hidden state vector \(h_t\), based on the current input \(x_t\) and the previous hidden state \(h_{t-1}\):

\begin{equation}
h_t = \sigma(W_h \cdot h_{t-1} + U_x \cdot x_t + b_h)
\end{equation}

Where:
\begin{align*}
h_t & \text{ is the hidden state vector at time } t. \\
\sigma & \text{ typically represents } \tanh \text{ or ReLU activation function.} \\
W_h & \text{ is the weight matrix for the hidden state.} \\
h_{t-1} & \text{ is the previous time step's hidden state.} \\
U_x & \text{ is the weight matrix for the input.} \\
x_t & \text{ is the input vector at time } t. \\
b_h & \text{ is the bias vector for the hidden state.}
\end{align*}

2. \textbf{Output}:

The output \(y_t\) at each time step is often based on the current hidden state:

\begin{equation}
y_t = V \cdot h_t + b_y
\end{equation}

Where:
\begin{align*}
y_t & \text{ is the output at time } t. \\
V & \text{ is the weight matrix for the output.} \\
h_t & \text{ is the current hidden state.} \\
b_y & \text{ is the bias vector for the output.}
\end{align*}

RNNs are adept at modeling temporal dependencies in sequential data, making them suitable for tasks like ECG signal analysis. However, they can be impacted by vanishing and exploding gradient problems, affecting their performance. RNNs generally have higher latency and throughput than CNNs.

\item \textbf{DBN (Deep Belief Network):} Deep Belief Networks (DBNs) are a class of artificial neural networks that have garnered attention for their remarkable capabilities in unsupervised learning and feature representation. Developed as a generative model, DBNs are composed of multiple layers of restricted Boltzmann machines (RBMs), and they have found applications in diverse domains such as image recognition, recommendation systems, and dimensionality reduction.

\section*{DBN Mathematical Model}

1. \textbf{Restricted Boltzmann Machine (RBM)}:

The fundamental building block of a DBN is an RBM, consisting of a visible layer (input data) and a hidden layer. The RBM's energy is defined as:

\begin{align}
E(v, h) = -\sum_{i} \sum_{j} w_{ij} v_i h_j - \sum_{i} a_i v_i - \sum_{j} b_j h_j
\end{align}

Where:
\begin{align*}
E(v, h) & \text{ is the energy for visible vector } v \text{ and hidden vector } h. \\
v_i, h_j & \text{ are the binary states of visible and hidden units, respectively.} \\
w_{ij} & \text{ is the weight between visible unit } v_i \text{ and hidden unit } h_j. \\
a_i, b_j & \text{ are biases for visible and hidden units, respectively.}
\end{align*}

2. \textbf{Joint Probability Distribution}:

The joint probability distribution is defined using the energy function:

\begin{equation}
P(v, h) = \frac{e^{-E(v, h)}}{Z}
\end{equation}

Where:
\begin{align*}
P(v, h) & \text{ is the joint probability distribution.} \\
Z & \text{ is the partition function, normalizing the distribution.}
\end{align*}

3. \textbf{Training}:

DBN training involves:
- \textbf{Pretraining}: Each RBM is trained layer-wise using unsupervised learning (e.g., Contrastive Divergence).
- \textbf{Fine-tuning}: The entire network is fine-tuned using supervised methods like backpropagation.

\begin{figure}[h]
    \graphicspath{ {D:\Stack} }
    \center \includegraphics[width=6cm, height=4cm]{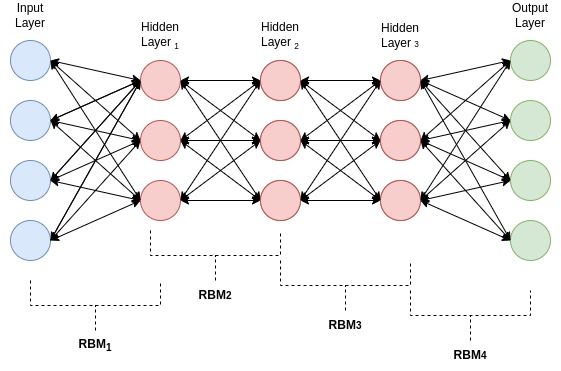}
    \caption{DBN Architecture}
    \label{fig:enter-label}
\end{figure}   
\cite{DBNarchitecture}

4. \textbf{Stacking RBMs}:

Trained RBMs are stacked to form a DBN, capturing increasingly abstract data features.

DBNs, with multiple layers of RBMs or autoencoders, excel in feature extraction and classification, but may not be as effective as LSTMs or CNNs for sequential data like ECG signals. Their performance varies based on architecture.

\end{itemize}
Based on the above comparison, CNNs are generally considered the best choice for ECG analysis when low latency and high throughput are essential, as shown in Table \ref{tab:comparison}. However, the specific choice of model depends on the requirements of the application and the nature of the ECG data. Experimenting with different models can be useful to determine which one performs best for a specific use case.

\begin{table}[h]
\centering
\caption{Resource utilization summary}
\label{table:utilization}
\begin{tabular}{|c|c|c|c|}
\hline
\multicolumn{4}{|c|}{Zynq®-7000 SoC} \\ \hline
\textbf{Resource} & \textbf{Utilization} & \textbf{Available} & \textbf{\% Utilization} \\ \hline
LUT & 17579 & 74000 & 23.75\% \\
FF & 20060 & 106400 & 18.85\% \\
BRAM & 1374 & 3300 & 41.64\% \\
IO & 36 & 150 & 24\% \\
DSP & 85 & 160 & 53.13\% \\ \hline
\end{tabular}
\end{table}

\begin{table*}
\centering
\caption{Comparison of LSTM, CNN, RNN, and DBN models based on various parameters}
\label{tab:comparison}
\begin{tabular}{@{}lcccccccc@{}}
\toprule
\textbf{Models} & \textbf{Accuracy} & \textbf{Precision} & \textbf{Recall} & \textbf{F1-score} & \textbf{Training time} & \textbf{Model complexity (params)} & \textbf{Throughput [GOP/s]} & \textbf{Latency} \\ \midrule
LSTM & 81\% & 28\% & 18\% & 16\% & 202.43 s & 29,477 & 2439.44 & 37 ms \\
CNN & 99\% & 20\% & 4\% & 8\% & 622.63 s & 3,245,637 & 2039.21 & 14 ms \\
RNN & 69\% & 18\% & 20\% & 19\% & 7.20 s & 276,737 & 378.10 & 43 ms \\
DBN & 19.43\% & 39\% & 20\% & 66\% & 304.18 s & 89,345 & 2237.40 & 30 ms \\ \bottomrule
\end{tabular}
\end{table*}

\subsubsection{CPU/GPU Implementation}
We implemented our model on both CPU and GPU, utilizing Python for execution. The computational power of NVIDIA's GeForce RTX 3060 GPU and Intel's Core i9 12900H CPU was leveraged, each optimized for different tasks, ensuring efficient execution.

\subsubsection{Hardware Accelerator MAC Components}
The MAC operations for the primary components of the CNN architecture are quantified as follows:

\begin{enumerate}
    \item \textbf{Convolutional Layers}: These layers involve convolution operations, which are essentially matrix multiplications between the input data and the convolutional filters or kernels. Let \(F\) be the number of filters, \(D\) the dimension of each filter, and \(I\) the size of the input feature map.
    \begin{equation}
        \text{MAC}_{\text{Conv}} = F \times D^2 \times I^2
    \end{equation}

    \item \textbf{Pooling Layers}: Pooling (e.g., max pooling) reduces the spatial dimensions of the feature maps. The MAC operations for pooling are typically less computationally intensive compared to convolutional layers. If \(P\) represents the size of the pooling window:
    \begin{equation}
        \text{MAC}_{\text{Pool}} = I^2 / P^2
    \end{equation}

    \item \textbf{Fully Connected Layers}: These layers are similar to traditional neural network layers and involve matrix multiplications between the flattened feature map and the layer's weights. Let \(C\) be the number of connections in a fully connected layer.
    \begin{equation}
        \text{MAC}_{\text{FC}} = C
    \end{equation}
\end{enumerate}

The total MAC operations for a CNN architecture are the sum of the MAC operations from its convolutional, pooling, and fully connected layers:
\begin{equation}
    \text{MAC}_{\text{Total}} = \text{MAC}_{\text{Conv}} + \text{MAC}_{\text{Pool}} + \text{MAC}_{\text{FC}}
\end{equation}

\subsubsection{Throughput Analysis}
Throughput is computed using MAC counts and latency:
\begin{equation}
\text{Simulation Time} = \frac{\text{Total Inference Time}}{\text{Total number of inference samples}}
\end{equation}
\begin{equation}
    \text{Throughput} = \frac{\# \text{MACs}}{\text{Latency (Simulation Time)}}
\end{equation}

\subsection{FPGA Implementation with Tensil's Open Source Inference Accelerator}
We provide a guide on implementing a ResNet-20 convolution model, trained on the MIT-BIH Arrhythmia Database, on a PYNQ Z1 FPGA using Tensil's open-source inference accelerator (Fig. \ref{Figure 3}). Resource utilization on a Zynq®-7000 SoC is detailed in Table \ref{table:utilization}.

\begin{figure*}
    \graphicspath{ {D:\Stack} }
    \center \includegraphics[width=0.6\textwidth]{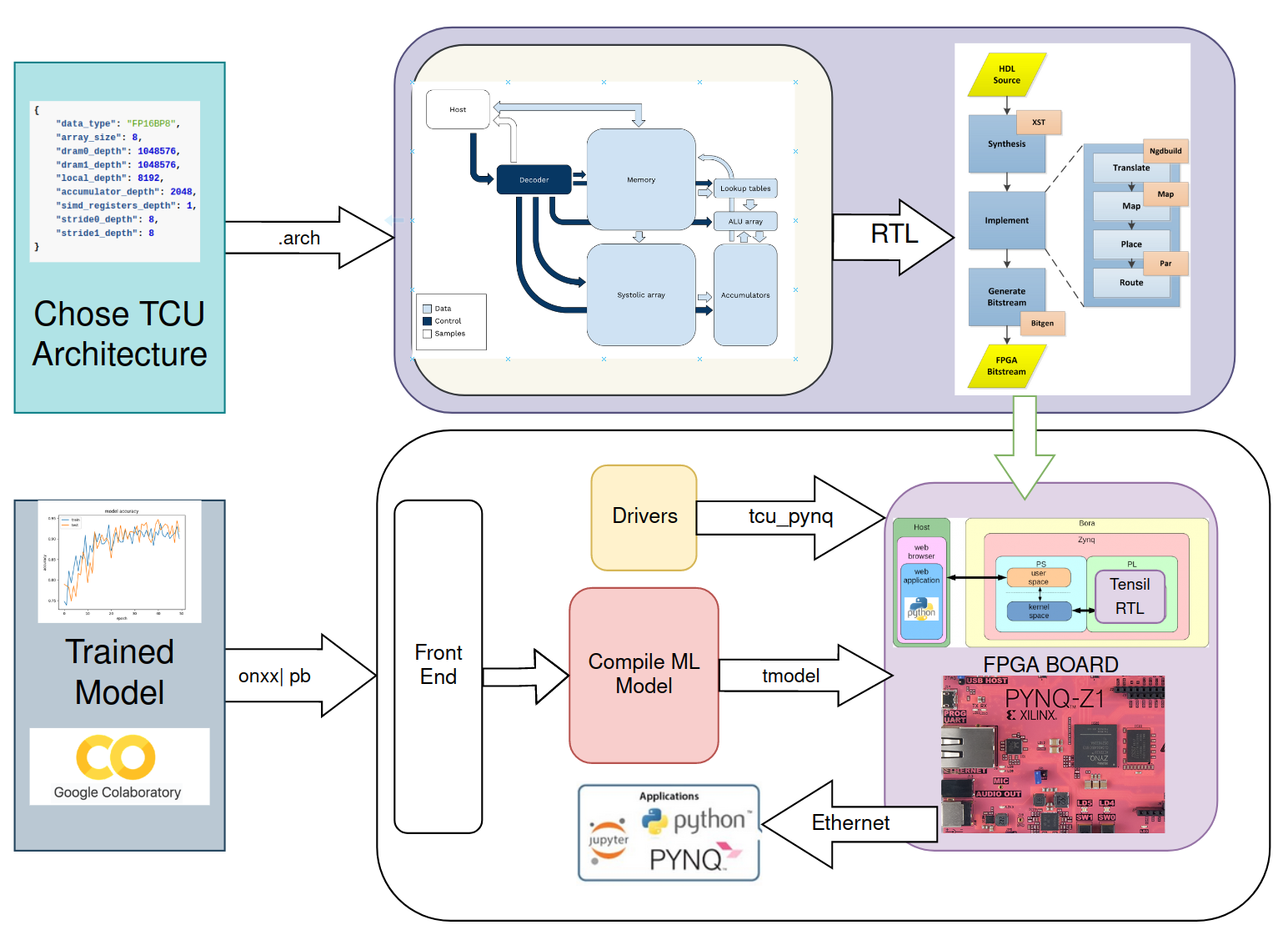}
    \caption{Block Diagram of Implementation}
    \label{Figure 3}
\end{figure*}

This section presents a detailed guide on implementing a ResNet-20 convolution model, trained not on CIFAR but the MIT-BIH Arrhythmia Database, on a PYNQ Z1 FPGA using Tensil's open-source inference accelerator, as depicted in Fig. \ref{Figure 3}. Table \ref{table:utilization} provided indicates the resource utilization on a Zynq®-7000 SoC when running the implemented model. It uses 23.75\% of available Lookup Tables (LUT), 18.85\% of Flip-Flops (FF), and 41.64\% of Block RAM (BRAM). The model also employs 24\% of available Input/Output (IO) resources and 53.13\% of Digital Signal Processors (DSP). This summary highlights the efficient use of the Zynq®-7000 SoC's resources by the model, ensuring it operates effectively within the constraints of the system.

\subsubsection{Installation and Setup of the Tensil Toolchain}

The Tensil toolchain is a suite of tools designed to facilitate FPGA development, specifically for running machine learning models on FPGA. Docker is used to host the Tensil toolchain, allowing for simple installation and setup. For those unfamiliar with Docker, it's a platform that enables developers to package and distribute their applications in a manner that is platform-independent. It's essential to ensure Docker is installed before proceeding with the Tensil toolchain setup.

\subsubsection{Selection of Architecture}

The choice of architecture depends on the specific requirements of your machine learning model and the resources available on your FPGA. It's crucial to consider factors such as the complexity of your model, the amount of data you'll be processing, and the computational resources of your FPGA.

\subsubsection{TCU Accelerator Design and Synthesis for PYNQ Z1}

In this section, a custom TCU (Tensor Compute Unit) accelerator design is generated and synthesized for the PYNQ Z1 board using the Tensil toolchain. The goal here is to create a hardware design that can efficiently execute the machine learning model on the FPGA.

\subsubsection{PS-PL Configuration and Smart Interconnect}

The next steps involve setting up the configuration between the Processing System (PS) and Programmable Logic (PL) and establishing a smart interconnect. These steps ensure efficient data transfer and proper interfacing between the various components of the design.

\subsubsection{Compilation of the ML Model}

To execute the machine learning model on the FPGA, we compile the model into a ".tmodel" file. The ".tmodel" file contains the model's structure and parameters in a format that can be executed on the FPGA. Additional files ".tprog" and ".tdata" are also generated, providing instructions and data for the FPGA execution.

\subsubsection{Execution Using PYNQ}

In this stage, all the previously generated and compiled files are deployed onto the PYNQ environment for execution. A PYNQ environment refers to a Python-based ecosystem that simplifies the usage and programming of Xilinx Zynq SoCs. It's important to set up this environment correctly on the FPGA for successful execution. An FPGA image is essentially a binary file that contains the configuration data for the FPGA. This image is crucial for defining the functionality of the FPGA hardware during runtime.

\section{Results}

 \begin{figure}[h]
        \centerline{\includegraphics[width = 0.99\columnwidth]{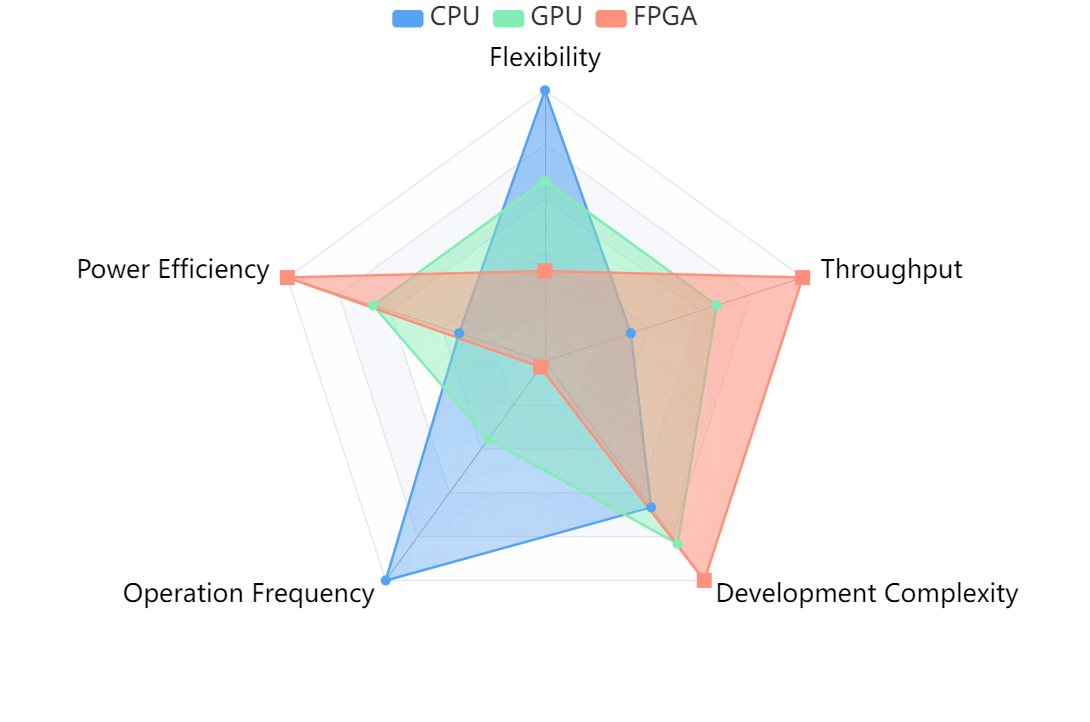}}
        \caption{Our Framework on Different Heterogeneous Devices.}
        \label{Chart1}
    \end{figure}

The chart illustrates that CPUs, with their general-purpose design, offer flexibility but often at the cost of power efficiency. GPUs, with their many-core architecture, provide higher throughput for parallel computing tasks but lag in power efficiency compared to FPGAs. FPGAs, with their reconfigurable nature, achieve superior power efficiency and throughput for specific applications but require expertise in digital design.

\begin{table*}
    \renewcommand{\arraystretch}{1.2}
    \setlength{\tabcolsep}{3pt}
    \centering
    \caption{Comparisons with previous implementations.}
    \label{table:table_3}
    \resizebox{0.8\linewidth}{!}{
    \begin{tabular}{|c|c|c|c|c|c|c|}
    \hline
     & \cite{zhao201913} & \cite{xia2019novel} & \cite{wang2019energy} & \cite{wong2022energy} & \cite{degirmenci2022arrhythmic} & Our \\
    \hline
    Convolution Type & 1-D & 1-D & 1-D & 2-D & 2-D & 2-D \\
    \hline
    Platform & FPGA Pynq-Z2 & CPU-i7 & - & iCE40UP5k & GPU RTX 2080 Ti& FPGA Pynq-Z1 \\
    \hline
    No. Input Samples & 512 & 200 & 400 & 10x10 & 64x64 & 187 \\
    \hline
    Activation & - & ReLu & - & bTanH & ReLu & ReLu  \\
    \hline
    Num of MACs & 929,650 & 1,289,312 & 749,620 & 27,153 & 58.1 M & 47,560 \\
    \hline
    Clock & 25 MHz & 3.7 GHz & - & 100 MHz & 1350 MHz & 100 MHz  \\
    \hline
    Accuracy & 98.9 & 99.8 & 98.4 & 96.8 & 99.7 & 99.1  \\
    \hline
    Power & 13.34 $\mu$W & 84 W & 141 mW & 227.3 $\mu$W & 108 W & 1.53 W  \\
    \hline
    \end{tabular}}
\end{table*}

\section{Conclusions}

Our research demonstrates the effectiveness of Tensil AI's open-source inference accelerator in optimizing neural networks and implementing them on FPGAs for high-performance computing applications. We achieved significant results across CPUs, GPUs, and FPGAs, contributing to the growing body of work on FPGA-based NN inference systems. Our approach, leveraging 2-D convolution on a PYNQ Z1 FPGA platform with a ReLu activation function, processed 187 input samples with a high accuracy of 99.1\% and consumed only 1.53 W of power. These findings underline the potential of FPGAs in high-throughput, power-efficient computing applications. We plan to incorporate Dynamic Partial Reconfiguration (DPR) into our system to improve performance further and explore Tensil AI's flexibility in handling diverse machine learning models.

\section{Future Research Directions}

The findings of our current study lay the groundwork for several promising research trajectories:

\subsection{Integration of Dynamic Partial Reconfiguration (DPR)}
Future research will explore the integration of DPR into FPGA systems. This advancement aims to enhance the flexibility and performance of FPGAs by enabling on-the-fly reconfiguration capabilities. This aspect of research will investigate the potential improvements in computational efficiency and adaptability to varying workloads.

\subsection{Optimization of Data Processing Stages}
Optimizing the initial and post-data processing stages is another research avenue. Utilizing tools like Tensil AI, this endeavor will focus on refining the computational pipeline to enhance efficiency and reduce processing overheads, thereby streamlining the entire data flow from input to output.

\subsection{Extending Model Compatibility with Tensil AI}
Our research will also extend the scope of Tensil AI’s compatibility with a wider array of machine learning models. This will involve comprehensive testing and development to showcase the platform’s versatility across different machine learning paradigms and its applicability to diverse computational tasks.

\subsection{Benchmarking Against Emerging Hardware Technologies}
An important future direction is benchmarking FPGA performance against emerging hardware technologies. This comparative analysis will provide insights into the relative strengths and limitations of FPGAs in the context of advancing computational technologies.

\subsection{Studies on Energy Consumption and Sustainability}
In response to the growing emphasis on sustainable technology practices, in-depth studies focusing on the energy consumption patterns of FPGA implementations are planned. These studies will seek to develop strategies for energy-efficient operations and sustainable system designs.

\subsection{Real-world Deployment and Clinical Trial Collaborations}
Collaborating with medical institutions for the real-world deployment and clinical trials of FPGA-based systems, particularly focusing on real-time ECG signal analysis, will be a significant step towards practical application. This collaboration aims to validate the efficacy of our implementations in clinical settings.

\subsection{Enhancing the Usability of FPGA Development Toolchain}
Recognizing the complexity involved in FPGA development, future work will also involve efforts to enhance the usability of the FPGA development toolchain. This initiative aims to make FPGA technology more accessible to a broader range of practitioners, including those without extensive expertise in hardware design.

\subsection{Cross-Domain Application Exploration}
Exploring the adaptability of FPGA-based systems to other domains, such as autonomous systems and IoT devices, represents an exciting frontier. This will involve adapting the existing framework to meet the specific computational demands of different applications.

\subsection{Co-Development Practices for Hardware and Software}
Investigating co-development practices that concurrently enhance both the hardware layout and the software framework is a key research area. This integrated approach aims to synergize hardware and software development, maximizing the performance capabilities of FPGAs.

\subsection{Development of Advanced Compiler Strategies}
Lastly, the development of advanced compiler strategies to optimize the performance of neural networks on FPGAs will be a crucial focus. This will include exploring innovative forms of parallelism and custom instruction sets specifically tailored to machine learning tasks, potentially leading to significant advancements in FPGA-based machine learning implementations.

\printbibliography

\end{document}